\documentclass[a4paper,conference]{IEEEtran}
\ifCLASSINFOpdf
\usepackage[pdftex]{graphicx}
\DeclareGraphicsExtensions{.pdf,.jpeg,.png}
\else
\fi
\usepackage{amsmath, amssymb}
\usepackage{mathrsfs}
\usepackage{xcolor}
\usepackage{array}
\usepackage{multirow}
\usepackage{url}

\usepackage{booktabs} 
\usepackage{hyperref}

\begin{document}
%
\title{Transformers for 1D signals in Parkinson's disease detection from gait}

\author{\IEEEauthorblockN{Duc Minh Dimitri Nguyen, Mehdi Miah, Guillaume-Alexandre Bilodeau}
\IEEEauthorblockA{LITIV Laboratory, Dept. of Computer engineering and Software engineering\\Polytechnique Montréal\\
Montréal, Canada\\
duc.md.nguyen@ulb.be, \{mehdi.miah, gabilodeau\}@polymtl.ca}
\and
\IEEEauthorblockN{Wassim Bouachir}
\IEEEauthorblockA{TÉLUQ University\\
Department of Science and Technology\\
Montréal, Canada\\
wassim.bouachir@teluq.ca}}


\maketitle

\begin{abstract}
 This paper focuses on the detection of Parkinson's disease based on the analysis of a patient's gait. The growing popularity and success of Transformer networks in natural language processing and image recognition motivated us to develop a novel method for this problem based on an automatic features extraction via Transformers. The use of Transformers in 1D signal is not really widespread yet, but we show in this paper that they are effective in extracting relevant features from 1D signals. As Transformers require a lot of memory, we decoupled temporal and spatial information to make the model smaller. Our architecture used temporal Transformers, dimension reduction layers to reduce the dimension of the data, a spatial Transformer, two fully connected layers and an output layer for the final prediction.
 Our model outperforms the current state-of-the-art algorithm with 95.2\% accuracy in distinguishing a Parkinsonian patient from a healthy one on the Physionet dataset. A key learning from this work is that Transformers allow for greater stability in results. The source code and pre-trained models are released in \href{https://github.com/DucMinhDimitriNguyen/Transformers-for-1D-signals-in-Parkinson-s-disease-detection-from-gait.git}{https://github.com/DucMinhDimitriNguyen} \footnote{Code and pretrained models will be published on Github upon paper acceptance.}.
\end{abstract}


%
\IEEEpeerreviewmaketitle

\section{Introduction}

Parkinson's disease (PD) affects between 12 and 15 in every 100 000 inhabitants, making it the second most common neurological disorder after Alzheimer's disease~\cite{hirtz2007common}. Age is the main factor explaining the onset of the disease: its prevalence in industrialized countries among the people over 60 reaches 1\%~\cite{de2006epidemiology}. This disease reduces not only the life expectancy of patients~\cite{de2006epidemiology}, but is also an economic burden for society~\cite{poewe2017parkinson}. Currently, there is no remedy to cure people suffering from PD. An early detection of the first symptoms of the disease allows the administration of drugs to mitigate the long-term effects. However, diagnosing PD is a complex task due to inter-individual variability, leading to false diagnoses resulting from a lack of knowledge or subjective errors by physicians.

This disease is caused by a lack of dopamine, a chemical messenger in the brain, causing motor and non-motor symptoms. Among the former, static tremors, rigidity, slowness of movement and postural instability are usually observed in patients. Non-motor symptoms are also described such as sleep disorder, speech disturbance and a loss of smell~\cite{poewe2017parkinson}. 

Recently, some researchers have been developing automatic methods to diagnose PD. As impaired gait is one of the most common characteristic of PD, gait analysis is a non-invasive and inexpensive method to detect the disease. A Parkinsonian gait is distinguished by smaller steps, a slower gait cycle, a longer stance phase, and a flat foot strike instead of a toe-to-heel strike~\cite{morris2001biomechanics}. The methods developed are either based on 1) hand-crafted features, such as stance time, swing time, speed, step length or stride length~\cite{balaji2020supervised}, combined with classical machine learning methods, such as support vector machines, decision trees or k-nearest neighbors, or 2) based on end-to-end learning methods, such as deep neural networks. Indeed, since the breakthrough results of artificial neural networks in vision~\cite{krizhevsky2012imagenet}, deep learning has gradually been adapted to other fields such as natural language processing. The Transformer architecture~\cite{vaswani2017attention}, which was originally developed for textual data, was then adapted for vision tasks~\cite{dosovitskiy2021an, Liu_2021_ICCV}. It is an encoder-decoder framework where the inputs are one sequence of data and the outputs another sequence. This model is also based on an attention mechanism, that allows to consider the influence of each part of the sequence.

In this paper, we tackle the problem of PD detection using gait. This can be achieved with the help of foot sensors. To do so, 18 1-dimensional signals are collected from the walk of a patient. These signal vectors represent the vertical ground reaction force (VGRF) in function of the time captured by 18 foot sensors. 1D signals are first divided into segments. Then, our deep learning model, based on Transformers, classifies those segments into the corresponding category (Parkinsonian or non Parkinsonian). Finally, we perform a majority voting using all the segments of a walk to determine whether the patient should be classified as Parkinsonian or not. 

The main methodological novelty brought by our model comes from the use of Transformers as features extractors. Only the encoder part of the traditional Transformers used in natural language is used. The idea is that we can use this encoder to represent useful information, by exploiting their abilities to capture temporal and spatial dependencies. To limit the complexity, our model first uses a Transformer encoder to capture temporal dependencies, and then a second encoder to learn spatial dependencies between all foot sensors.

To summarize, our main contributions are as follows:
\begin{itemize}
    \item we present a new method to detect Parkinson's disease with a Transformer-based algorithm, that first applies temporal attention on separate sensor signals, followed by spatial attention to build multisensor spatio-temporal gait features; 
    \item our method is competitive and more stable compared to state-of-the-art works, with an accuracy of 95.2\% on the PhysioNet dataset and a lower variance with respect to recent methods.
\end{itemize}

\section{Related Work}

As Parkinson's disease has numerous symptoms, several methodologies have been developed to detect it. For example, patients are requested to draw spirals on a graphic tablet with a digital pen~\cite{canturk2021fuzzy, khatamino2018deep}. We can also mention the analysis of the speech to detect symptoms of PD~\cite{moro2019forced, upadhya2018thomson, zahid2020spectrogram}. 

In our work, we focus on PD detection from gait data, which consists of vertical ground reaction force (VGRF) signals captured using foot sensors. In this context, Ertugrul et al.~\cite{ertuugrul2016detection} proposed an algorithm based on shifted 1D local binary patterns (1D-LBP) and machine learning classifiers. They used 18 VGRF input signals coming from foot sensors of Parkinson's patients and control subjects. For each signal, they applied shifted 1D-LBP to construct 18 histograms of the 1D-LBP patterns, from which they extracted statistical features, such as entropy, energy and correlation. Finally, they concatenated the features from all the 18 histograms and used various supervised classifiers, such as random forest and multi-layer perceptron (MLP) to classify feature vectors. Balaji et al~\cite{balaji2020supervised} extracted statistical and kinematic features such as swing time, swing stance ratio, cadence, speed or step length. They fed these hand-crafted features into several machine learning techniques such as a decision tree, a support vector machine, an ensemble classifier and a Bayes classifier to assess the severity of the disease. Zhao et al~\cite{zhao2022severity} used an ensemble k-nearest neighbor on hand-crafted features to predict the severity of PD.

Since the revolution of deep learning in 2012, end-to-end learning algorithms have been used to detect PD. A comprehensive review on the use of neural networks for the detection of PD is available in the work of Alzubaidi et al~\cite{alzubaidi2021role}. Aversano et al~\cite{aversano2020early} employed a deep neural network directly on the data coming from the sensors. El Maachi et al. \cite{Elmaachi} presented a deep 1D convolutional neural network (1D-Convnet) for the Parkinson's disease detection and severity prediction from gait. Their model processed the 18 1D signals coming from foot sensors measuring the vertical ground reaction force. The first part of the network consists of 18 parallel 1D-Convnet to process each 1D signal. The second part is a fully connected network that connects the concatenated outputs of the 1D-Convnets to obtain a final classification. The model that will be presented in this paper was inspired by this last work, which currently holds the state-of-the-art (SOTA) accuracy in the classification of Parkinsonian patients based on gait analysis.

2D-Convnet-based models were developed by Hoang et al~\cite{hoang2019gait} by concatenating all signals into a two-dimensional image. Next, they used a 2D-Convnet to extract features from the image, which were then reshaped into a one-dimensional vector. A 1D-Convnet was finally used to capture the temporal effect for each segment of the walk. Pretrained 2D-Convnets were used in the work of Setiawan and Lin~\cite{setiawan2021implementation}. They converted the signal from 16 sensors into a spectogram image, which is a visualization commonly used to depict audio signals. Then, they used pre-trained models such as AlexNet, ResNet and GoogLeNet to assess the severity of the disease.

Convnets were also combined with Long-Short-Term Memory (LSTM)~\cite{hochreiter1997long}, a recurrent neural network architecture to extract features from the temporal and spatial domains. Zhao et al. \cite{Zhao} used a deep learning algorithm to detect Parkinson's disease. Their model was composed of a network to analyze the spatial distribution of forces with a 2D-Convnet and a second network to analyze the temporal distribution with a recurrent neural network. These two layers worked in parallel. The final classification was decided by the average of both output channels. Then, Xia et al~\cite{xia2019dual} improved the architecture by differentiating the left gait from the right with an attention-enhanced LSTM. After extracting representation with a 2D-Convnet, they constructed robust features for both feet. In addition to the network change, input sequences were based on a gait cycle instead of extracting segments of walk.

Previous works used recurrent neural networks, convolutional neural networks and other architectures that are now slowly being replaced by Transformers in different kinds of applications, such as image recognition as demonstrated in~\cite{dosovitskiy2021an}. Indeed, a Transformer encoder relies on an attention mechanism to weight the representations from each element of a sequence. These representations are then fused with a fully connected layer. The idea of this paper is to capitalize on the ability of Transformers to capture signal dependencies to improve the feature extraction part of the algorithm. By doing so, we show that we outperform the current state-of-the-art method \cite{Elmaachi}. 

\section{Proposed Transformer model}

\begin{figure*}[ht]
\centering
    \includegraphics[width=18cm, trim={0 80 70 0},clip]{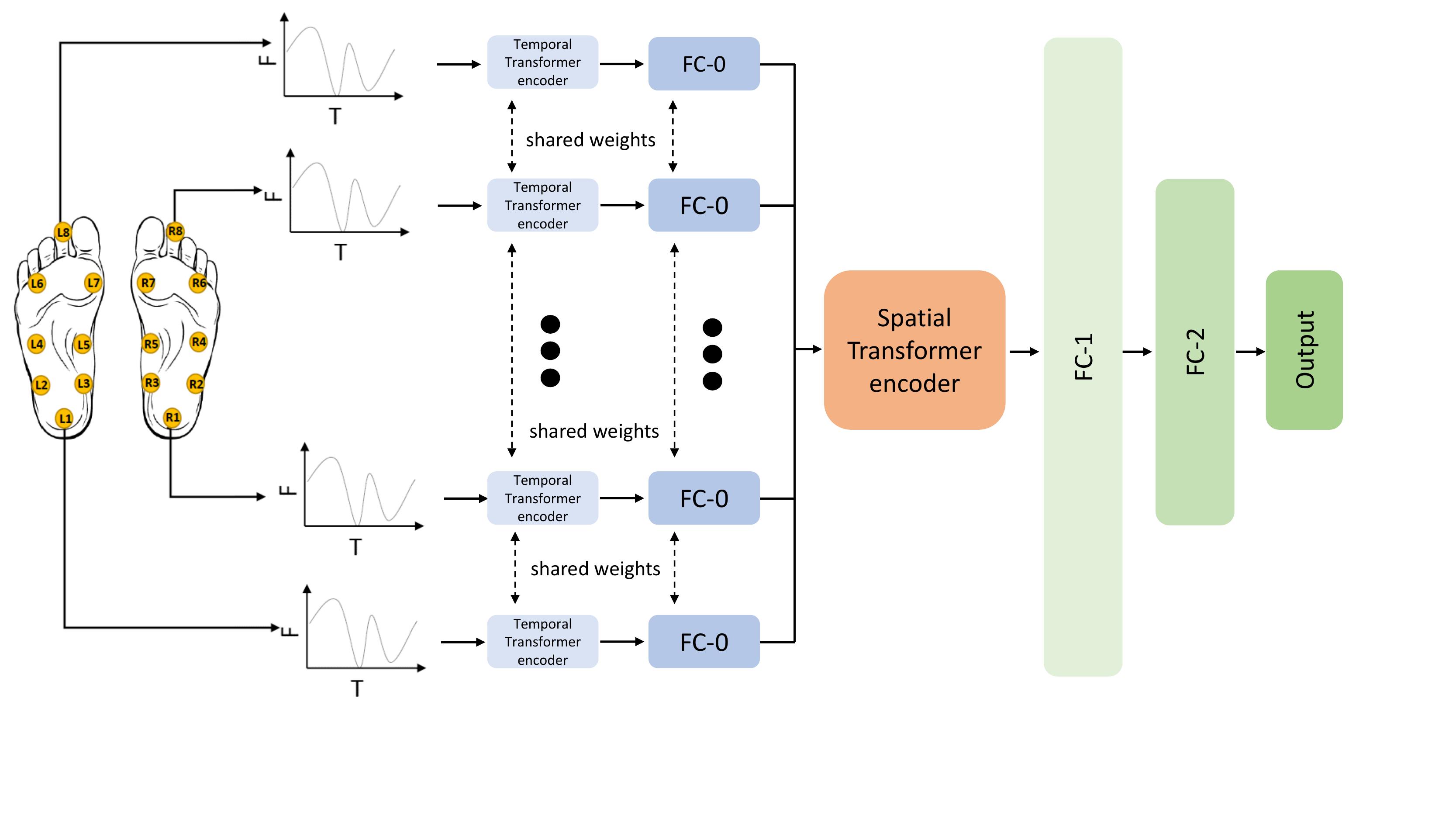}
    \caption{Architecture of our Transformer model. It is composed of a feature extractor (Temporal Transformer encoder, FC-0, Spatial Transformer encoder), and a classifier made of two fully connected layers and an output layer (FC-1, FC-2, Output). This view omits the positional encodings and the segmentations of the input sequence.}
    \label{model}
\end{figure*}

Our proposed model is illustrated in figure \ref{model}. It is composed of two main parts: 1) a feature extractor made of Transformers (Temporal Transformer encoder, FC-0, Spatial Transformer encoder), and 2) a classifier made of two fully connected layers and an output layer (FC-1, FC-2, Output). The first part is where we are contributing, by introducing a novel feature extractor using Transformers. The second part corresponds to fully connected layers using the features extracted as input, to output the final classification. 

The idea behind the automatic feature extractor comes from two key observations. Firstly, through several experiments, we observed that temporal and spatial dependencies are important for the model to correctly classify the patient. Indeed, the Transformers can be used to capture temporal dependencies of each sensor, which correspond to the link between two values of a vector separated by a certain amount of time. Furthermore, the Transformers can also be used to capture the spatial dependencies between each set of vectors coming from the 18-foot sensors. Each foot sensor is placed in a different position on the foot, which can be useful information for the algorithm. Secondly, Transformers are very memory consuming. To make our method more easily applicable and smaller, we decided to use the Transformers in two stages to capture temporal dependencies at first, and then to capture spatial dependencies, instead of capturing both with a single Transformer. Therefore, after capturing temporal dependencies, a dimension reduction is performed using a fully connected layer. This allowed us to then use a Transformer with less data to capture of the spatial dependencies. The proposed method is detailed in the following subsections.

\subsection{Data preprocessing}

\begin{figure}[ht]
\centering
    \includegraphics[width=8cm, trim={0 140 450 0}, clip]{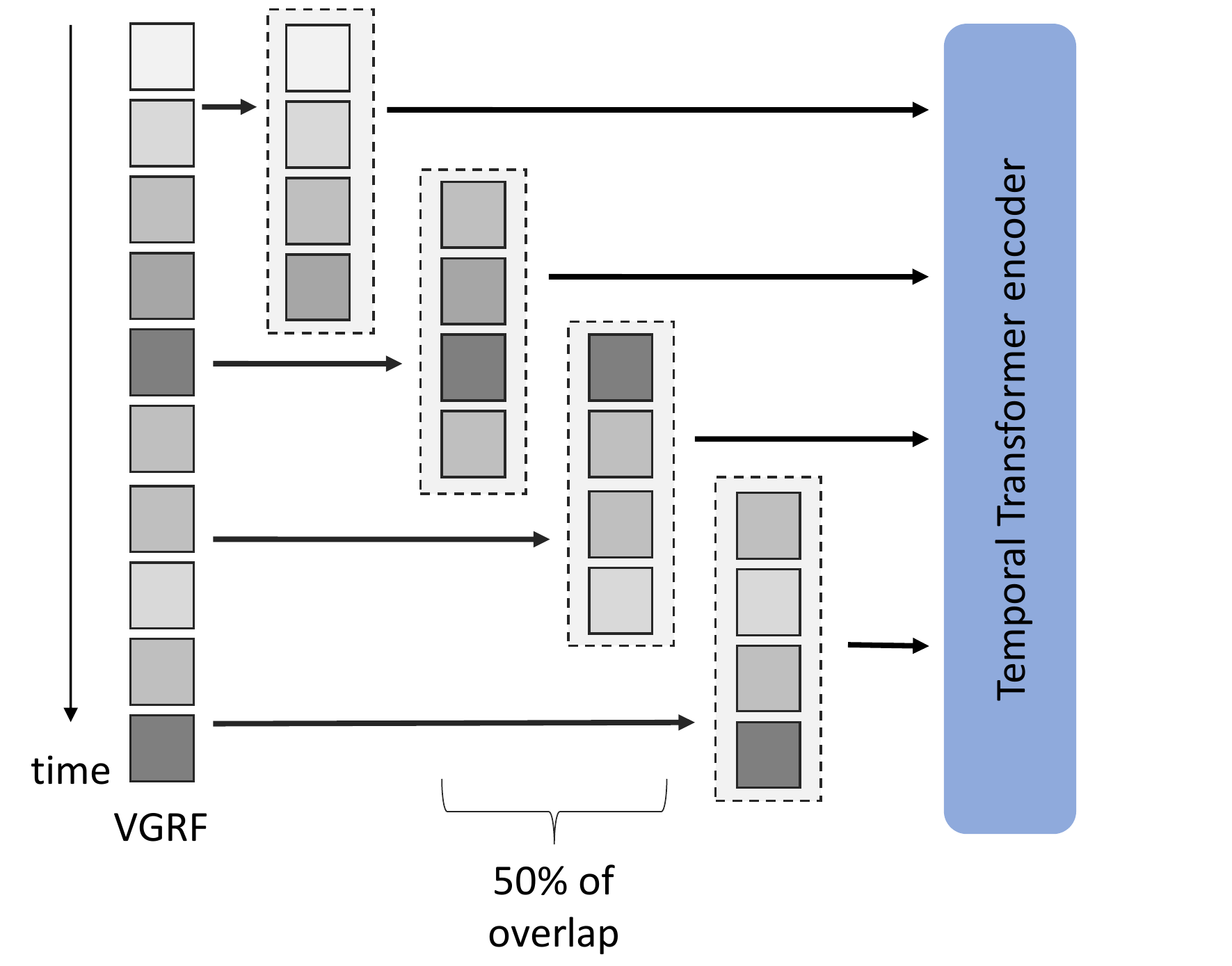}
    \caption{Segmentation: from a sequence, we obtain multiple fixed-sized sub-sequences which are the inputs of our model.}
    \label{segmentation}
\end{figure}

In order to have more data, each walk is divided into smaller segments of 100 time steps with 50\% overlap (the final dataset contains 64468 segments) as illustrated in figure \ref{segmentation}. In addition to providing more data, this segmentation of the walks allows us to keep our model small, as temporal Transformers have fewer parameters. Furthermore, this allows us to classify each walk using the combined classification of each segment.

The 100-time step has been chosen in a manner that enough information is stored in each segment, while keeping the vectors small enough, so that the Transformers can still be used. Indeed, too large vectors cannot be used due to the memory limitation.

\subsection{Temporal Transformer encoders}

Each temporal Transformer encoder block is composed of a multi-head attention and a feed forward network as proposed in BERT~\cite{vaswani2017attention} for natural language processing. One subtlety resides in the choice of the positional encoding. Because Transformers do not have a mechanism to take into account the ordering of a sequence, we have to add to the input data a positional encoding that can encode this ordering, as suggested in \cite{vaswani2017attention}. 

 Multiple positional encoding scheme are possible (learned and fixed) as shown in \cite{Gehring}. Here, we chose a fixed one with constant step in between the different positional encoding. This is because in our case, we separated the vectors in a fix and constant way. Each vector contains 100 elements. Unlike in natural language processing where a sentence may have various lengths, here the length of our Transformer input is known and fixed, allowing us to use the positional encoding described below. Our choice is inspired by what has been done in image recognition in~\cite{dosovitskiy2021an}.

\begin{figure}[ht]
\centering
    \includegraphics[width=8.8cm, trim={0 0 310 0}, clip]{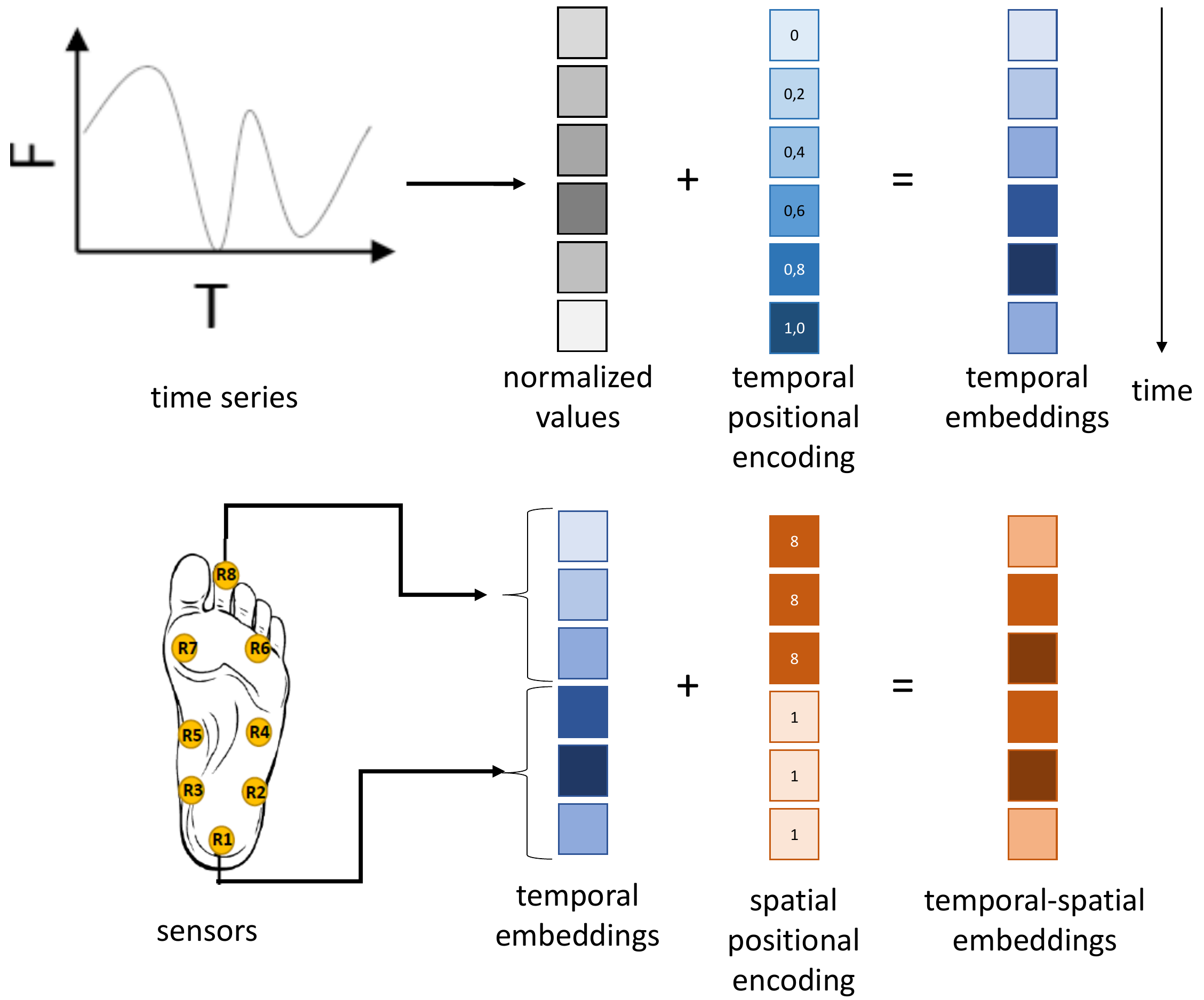}
    \caption{Positional encoding: the spatio-temporal information is decoupled so that the model treats firstly temporal information then spatial information. Top: positional encoding for the temporal Transformer encoder, bottom: positional encoding for the spatial Transformer encoder.}
    \label{positional_encoding}
\end{figure}

The first 18 Transformers used to capture the temporal dependencies utilize the same sequence of number going from 0 to the size the vector (100 in our case) as the positional encoding. The positional encoding has been normalized (no element is exceeding 1) in order to not mask all the information present in the original vector when adding the positional encoding to it. The top part of figure~\ref{positional_encoding} illustrates how this has been implemented.

\subsection{Spatial Transformer encoder}

Hence, temporal dependencies are first retrieved through our temporal Transformer encoders. Then a dimension reduction is performed, followed by a concatenation of all of the reduced vectors to be used as input to the spatial Transformer encoder. The role of the spatial Transformer is to find dependencies between sensors.

Another positional encoding is added to the input of the spatial Transformer encoder. Here the input of the spatial Transformer consists of 18 outputs of the different temporal Transformer encoder that have gone through a dimension reduction. Each of the output of the different temporal Transformer encoders are reduced from 100 elements to 10 elements. Then, the 18 vectors of 10 elements are concatenated together to be fed to the spatial Transformer encoder. To take into account the fact that those 18 vectors come from 18 different foot sensors, the positional encoding corresponds to a shift of every element of a vector by the same constant. The value of this constant is between 0 and 17, normalized depending on the foot sensors the vector comes from. The bottom part of figure~\ref{positional_encoding} shows an example of how it has been implemented.

\subsection{Walk classification}

After encoding spatial dependencies, features are passed through two fully connected layers and the output layer for classification. Each segment will be classified and at the end, a majority vote will be performed to determine if the patient is Parkinsonian or not. 

\section{Experiments}

\subsection{Implementation details}

Table~\ref{hyper} gives the hyperparameters of our proposed method. In table \ref{hyper}, the number of blocks represents the number of encoder blocks for the Transformers. The multi-head attention has been used with an embedding dimension of 100 and 180, respectively for the temporal and spatial Transformer encoders. The concatenated layer represents the concatenation of the 18 vectors coming from the foot sensor, each being of size 10 due to the dimension reduction. 

Every fully connected layer has been used with the \textit{Selu} activation function (scaled exponential linear units)~\cite{klambauer2017self} except for the output where we have used a sigmoid activation function. 
Note that we obtained slightly better results when using \textit{Selu} instead of \textit{Relu} as the activation function. The learning rate used is 0.001.
100 epochs has been used with an early stopping. The early stopping has been monitored using the validation loss with a minimum change of 0.01 in the monitored quantity to qualify as an improvement. Furthermore, 20 epochs with no improvement will stop the training.
The batch size used is 110.



\begin{table}[ht]
\setlength\extrarowheight{3pt}
\caption{Values of the hyperparameters}
\centering
\begin{tabular}{|c|c|c|c|c|}
\hline
 & \begin{tabular}{@{}c@{}}\tiny{Nb}\\ \tiny{blocks} \end{tabular} & \tiny{Layer type}& \begin{tabular}{@{}c@{}}\tiny{Nb}\\ \tiny{units} \end{tabular} & \tiny{Dropout} \\ [0.5ex]
\hline
\begin{tabular}{@{}c@{}}Temporal\\ Transformer x18 \end{tabular} & 2 & Normalization & - & - \\
\cline{3-5}\cline{5-5}
 && \begin{tabular}{@{}c@{}}Multi-head\\ attention \end{tabular} &2&-\\
\cline{3-5}\cline{5-5}
& & Normalization &-& -\\
\cline{3-5}\cline{5-5}
&& FC &100& 0.1\\
\hline
\begin{tabular}{@{}c@{}}Stack temporal\\ Transformer \end{tabular}
&-&FC-0 &10&0.1\\
\cline{2-5}\cline{5-5}
&-& Concatenate &-&-\\

\cline{1-5}
\begin{tabular}{@{}c@{}} Spatial\\ Transformer \end{tabular} & 2 & Normalization & - & -\\
\cline{3-5}\cline{5-5}
 && \begin{tabular}{@{}c@{}} Multi-head\\ attention \end{tabular} &2&-\\
\cline{3-5}\cline{5-5}
& & Normalization &-& -\\
\cline{3-5}\cline{5-5}
&& FC &180& 0.1\\
\hline

FC & - & FC-1 & 100 & 0.1 \\
\cline{3-5}\cline{5-5}
 && FC-2 &20&0.1\\
 \cline{3-5}\cline{5-5}
 && Output &1&-\\

\hline
\end{tabular}

\label{hyper}
\end{table}

\subsection{Dataset and evaluation metrics}

We used the public dataset collected by Physionet~\cite{goldberger2000physiobank} \footnote{\url{https://physionet.org/content/gaitpdb/1.0.0/}}, which includes data reported by Frenkel-Toledo et al. \cite{frenkel2005effect,frenkel2005treadmill}, Yogev et al. \cite{yogev2005dual} and Hausdorff et al. \cite{hausdorff2007rhythmic}. In this dataset, PD patients and healthy controls are requested to walk with sensors tied to their shoes during two minutes as they walked at their usual, self-selected pace (from \cite{frenkel2005treadmill} and \cite{hausdorff2007rhythmic})
Furthermore, measures (from \cite{yogev2005dual}) recorded as subjects performed a second task while walking is also included in the dataset. In total, 306 walks have been recorded from 166 individuals. In fact, 93 (56\%) of the subjects are PD patients and 73 (44\%) healthy ones while 214 (70 \%) of the recorded walks are Parkinsonian and 92 (30\%) control walks. Because of the larger number of experiments performed on PD patients, more data have been collected for those type of patients resulting in an unbalanced dataset (70\% Parkinson walk and 30\% control walks). 
For each walk, 18 time series signals are available: 16 (8x2) VGRF recorded from 8 sensors on each foot and 2 total VGRFs under each foot. Table~\ref{dataset} contains some demographic statistics about the PD patients and the healthy group and table~\ref{walk} contains some information about the Parkinsonian walks and the control walks.

\begin{table}[ht]
\caption{Statistics about the patients of the Physionet dataset}
\centering
{\begin{tabular}{lcccccccc}
\toprule
Groups & Total subjects & \multicolumn{2}{c}{Gender} & & \multicolumn{4}{c}{Age (years)} \\
\cmidrule{3-4}
\cmidrule{6-9}
& & Male & Female & & -50 & 50-70 & +70 & Range \\
\midrule
PD patients & 93 & 58 & 35 & & 1 & 59 & 33 & 36-84 \\
Healthy     & 73 & 40 & 33 & & 1 & 56 & 16 & 20-77 \\
\bottomrule
\end{tabular}}
\label{dataset}
\end{table}

\begin{table}[ht]
\caption{Number of walks of the Physionet dataset for each studies they are originated from}
\centering
{\begin{tabular}{lcccccccc}
\toprule
Groups & Total walks & \multicolumn{2}{c}{Normal walk} &  & \multicolumn{4}{c}{Dual task walk} \\
\cmidrule{3-4}
\cmidrule{6-9}
& & \cite{frenkel2005treadmill} & \cite{hausdorff2007rhythmic} & & & & \cite{yogev2005dual}   \\
\midrule
Parkinsonian walk & 214 & 35 & 104 & & & & 75   \\
Control walk     & 92 & 29 & 25 & & & & 38   \\
\bottomrule
\end{tabular}}
\label{walk}
\end{table}

To assess the model, we used cross validation with 10 folds, the same folds as in \cite{Elmaachi}. Each Parkinson and control groups have been divided into 10 folds at the subject level to keep the same dataset balance (70\% Parkinson and 30\% control) for each fold. The division of each walk into smaller segments (100 time step with 50\% overlap) has been done inside each fold. Each segment was labeled with the subject category for the training. The model was trained to classify these segments and the final result is obtained through a majority vote. 

The control group is identified as the negative (N) group and the Parkinson group is the positive (P) group. Three metrics are used to measure the performance of our model : 
\begin{equation}
    Se = \frac{TP}{TP + FN}
\end{equation}
\begin{equation}
    Sp = \frac{TN}{TN+FP}
\end{equation}
\begin{equation}
    Acc = \frac{TP+TN}{TP+TN+FP+FN}
\end{equation}

Where : 
\begin{itemize}
    \item [$\bullet$] $Se$ is the sensitivity;
    \item [$\bullet$] $Sp$ is the specificity;
    \item [$\bullet$] $Acc$ is the accuracy;
    \item [$\bullet$] $TP$ ($TN$) the number of true positive (negative);
    \item [$\bullet$] $FP$ ($FN$) the number of false positive (negative).
\end{itemize} 

\subsection{Results}
Table \ref{results} presents the performance obtained by our model with respect to other related works. $SD$ represents the standard deviation obtained over the 10 folds. To have a reliable comparison and a similar experimental setup, we retrained the model of \cite{Elmaachi} using the code they provided. 

Our model outperforms the current SOTA method in the sensibility ($Se$) and the final accuracy ($Acc$). The specificity ($Sp$) is slightly lower than the deep neural network of \cite{Elmaachi}, but we do obtain a lower standard deviation. A key difference we can observe with the introduction of Transformers in our model is the gain in stability across the different folds. This is reflected in a lower standard deviation in all three metrics used to measure the performance of the algorithms.

This higher stability comes with a cost, as the training time of our model is about 4 times longer than the DNN of \cite{Elmaachi}. The Transformer architecture is indeed very memory consuming and needs a longer time to train. 

\begin{table}[ht]
\caption{Performance of our proposed method compared to SOTA. \textbf{Boldface} indicates the best method for each metric. \dag: results obtained by running their provided code.}
\centering
{\begin{tabular}{lccc}
\toprule
Methods & $Se\pm SD\%$ & $Sp\pm SD\%$& $Acc\pm SD\%$\\
\midrule
Transformer (ours) & $\textbf{98.1}\pm \textbf{3.2}$ & $86.8\pm \textbf{8.2}$& $\textbf{95.2}\pm\textbf{2.3}$ \\
DNN \cite{Elmaachi}\dag &   $97.0\pm4$ & $\textbf{88.5}\pm 11.3$ & $94.5\pm5.2$  \\
DNN \cite{Zhao} & $96.2\pm 3.8$ &  $76.7\pm \textbf{8.2}$&  $90.3\pm 2.9$  \\
MLP \cite{ertuugrul2016detection} &  $88.9$&  $82.2$&  $88.9$ \\
Random forest \cite{ertuugrul2016detection} &  n/a&  n/a&  86.9  \\
Naive Bayes \cite{ertuugrul2016detection} &   n/a & n/a &  76.1 \\
\bottomrule
\end{tabular}}
\label{results}
\end{table}

\subsection{Discussion}
To have more insights to understand the results, let us recall that the results obtained here come from a majority vote. Indeed, our model was trained to classify the segments of a walk. In fact, the accuracy obtained for the classification of the 64468 segments is 89\%. The majority vote allowed us to have a final accuracy of 95.2\% for the classification of a patient based on his/her entire walk. Hence, we can assume that the 11\% of misclassified segments are distributed across the patients and are not concentrated on a specific walk.
However, we can notice that the specificity is lower than the sensitivity. We can conclude that the majority of the misclassified segments comes from a control walk that has been detected as a Parkinsonian walk (false positive). In fact, certain healthy patients could have an atypical walk that looks like a Parkinsonian walk, which is misclassified by our algorithm. For this to happen, more than 50\% of the segments of that walk should be misclassified, resulting in a lower specificity. Note that this difficulty seems to be consistent across all the different SOTA method. This could be explained by the fact that the dataset is not very large.

Nevertheless, the sensitivity is very high, meaning that Parkinsonian patients are nearly always well detected. 

\subsection{Ablation study}

By removing elements of the final architecture, we can see how different elements of our model help to improve the final accuracy. Different models have been studied based on this principle. Table \ref{ablation} shows how each component of our model perform compared to the final architecture chosen. In model A, we removed all the temporal Transformers and replaced them by 1 dimensional convolutional networks. We also removed the dimension reduction layers (FC-0) and the spatial transformer. The second part (green part in figure 1) remains the same. In model B, we removed the dimension reduction layers (FC-0) and the spatial transformer. The temporal transformers and the second part of the architecture remain identical. In model C, we replaced the whole feature extractor by a single spatio-temporal Transformer that is now fed with the 18 1D signals simultaneously. The second part of the architecture is the same. Because of memory limitations we had to use smaller vector size in this case (vector of 50 elements, that is, the spatio-temporal Transformer is fed with a matrix of $18 \times 50$ elements). Finally, the final model corresponds to the one discussed in this article and presented in figure 1.

\begin{table}[ht]
\caption{Performance obtained in the ablation study. \textbf{Boldface} indicates best results.}
\centering
{\begin{tabular}{lccc}
\toprule
 & $Se\pm SD\%$ & $Sp\pm SD\%$& $Acc\pm SD\%$\\
\midrule
Model A & $97.0\pm 4.0$ & $\textbf{88.5}\pm 11.3$& $94.5\pm5.2$ \\
Model B & $95.8\pm 4.0$ & $81.3\pm 10.75$ & $91.4\pm4.4$  \\
Model C & $96.6\pm 4.7$ & $81.3\pm 18.6$ &  $92.0\pm 8.6$  \\
\midrule
Final model  &  $\textbf{98.1}\pm \textbf{3.2}$&  $86.8\pm \textbf{8.2}$&  $\textbf{95.2}\pm\textbf{2.3} $ \\
\bottomrule
\end{tabular}}
\label{ablation}
\end{table}

As we can observe, using the temporal Transformers (model B) instead of the 1-dimensional convolutional networks (model A) helps to decrease the standard deviation. However, the results are not as good as our final model, since spatial dependencies are not captured. By only exploiting one spatio-temporal Transformer (model C), the final accuracy obtained is slightly better than what we obtained with model B, but because of the large number of parameters necessary to attend on signal elements, only shorter time windows can be analyzed, which may not allow to capture well the information about the gait. Our combination of two Transformers allows to capture better gait information, while at the same time keeping memory demand reasonable. This was also observed in the context of video understanding by Bertasius et al.~\cite{bertasius2021SpaceTimeAttention} where it was shown that separating the spatial and temporal attention allows to capture long-range dependencies with a scalable design.

In the end, we conclude from this ablation study that removing or replacing components of our model results in decreasing performance.

\section{Conclusion}
In this paper, we presented a new approach for exploiting Transformer networks, in order to extract relevant gait features and detect Parkinson's disease from gait. The extracted features are then used in a classical feed forward network to output the classification result. 

Transformers are becoming more and more popular, especially in natural language and image processing. The goal of this work was to assess how this architecture can be used with 1D signals to classify gaits. As done in \cite{dosovitskiy2021an} for images, we were able to only use the encoder part to extract relevant features in the 1D signals. Another advantage of our model is that we can use intuitive positional encoding thanks to the non-variable length of the vectors.

A big challenge that is currently being tackled with Transformers is the memory consumption required by this type of architecture. Here, we proposed a way to exploit Transformer with a limited amount of memory, by splitting the temporal and spatial attention. This allowed us to achieve SOTA results.

\bibliographystyle{IEEEtran}
\bibliography{latex12}

\begin{thebibliography}{10}
\providecommand{\url}[1]{#1}
\csname url@samestyle\endcsname
\providecommand{\newblock}{\relax}
\providecommand{\bibinfo}[2]{#2}
\providecommand{\BIBentrySTDinterwordspacing}{\spaceskip=0pt\relax}
\providecommand{\BIBentryALTinterwordstretchfactor}{4}
\providecommand{\BIBentryALTinterwordspacing}{\spaceskip=\fontdimen2\font plus
\BIBentryALTinterwordstretchfactor\fontdimen3\font minus
  \fontdimen4\font\relax}
\providecommand{\BIBforeignlanguage}[2]{{%
\expandafter\ifx\csname l@#1\endcsname\relax
\typeout{** WARNING: IEEEtran.bst: No hyphenation pattern has been}%
\typeout{** loaded for the language `#1'. Using the pattern for}%
\typeout{** the default language instead.}%
\else
\language=\csname l@#1\endcsname
\fi
#2}}
\providecommand{\BIBdecl}{\relax}
\BIBdecl

\bibitem{hirtz2007common}
D.~Hirtz, D.~J. Thurman, K.~Gwinn-Hardy, M.~Mohamed, A.~Chaudhuri, and
  R.~Zalutsky, ``How common are the “common” neurologic disorders?''
  \emph{Neurology}, vol.~68, no.~5, pp. 326--337, 2007.

\bibitem{de2006epidemiology}
L.~M. De~Lau and M.~M. Breteler, ``Epidemiology of parkinson's disease,''
  \emph{The Lancet Neurology}, vol.~5, no.~6, pp. 525--535, 2006.

\bibitem{poewe2017parkinson}
W.~Poewe, K.~Seppi, C.~M. Tanner, G.~M. Halliday, P.~Brundin, J.~Volkmann,
  A.-E. Schrag, and A.~E. Lang, ``Parkinson disease,'' \emph{Nature reviews
  Disease primers}, vol.~3, no.~1, pp. 1--21, 2017.

\bibitem{morris2001biomechanics}
M.~E. Morris, F.~Huxham, J.~McGinley, K.~Dodd, and R.~Iansek, ``The
  biomechanics and motor control of gait in parkinson disease,'' \emph{Clinical
  biomechanics}, vol.~16, no.~6, pp. 459--470, 2001.

\bibitem{balaji2020supervised}
E.~Balaji, D.~Brindha, and R.~Balakrishnan, ``Supervised machine learning based
  gait classification system for early detection and stage classification of
  parkinson’s disease,'' \emph{Applied Soft Computing}, vol.~94, p. 106494,
  2020.

\bibitem{krizhevsky2012imagenet}
A.~Krizhevsky, I.~Sutskever, and G.~E. Hinton, ``Imagenet classification with
  deep convolutional neural networks,'' \emph{Advances in neural information
  processing systems}, vol.~25, pp. 1097--1105, 2012.

\bibitem{vaswani2017attention}
A.~Vaswani, N.~Shazeer, N.~Parmar, J.~Uszkoreit, L.~Jones, A.~N. Gomez,
  {\L}.~Kaiser, and I.~Polosukhin, ``Attention is all you need,'' in
  \emph{Advances in neural information processing systems}, 2017, pp.
  5998--6008.

\bibitem{dosovitskiy2021an}
\BIBentryALTinterwordspacing
A.~Dosovitskiy, L.~Beyer, A.~Kolesnikov, D.~Weissenborn, X.~Zhai,
  T.~Unterthiner, M.~Dehghani, M.~Minderer, G.~Heigold, S.~Gelly, J.~Uszkoreit,
  and N.~Houlsby, ``An image is worth 16x16 words: Transformers for image
  recognition at scale,'' in \emph{International Conference on Learning
  Representations}, 2021. [Online]. Available:
  \url{https://openreview.net/forum?id=YicbFdNTTy}
\BIBentrySTDinterwordspacing

\bibitem{Liu_2021_ICCV}
Z.~Liu, Y.~Lin, Y.~Cao, H.~Hu, Y.~Wei, Z.~Zhang, S.~Lin, and B.~Guo, ``Swin
  transformer: Hierarchical vision transformer using shifted windows,'' in
  \emph{Proceedings of the IEEE/CVF International Conference on Computer Vision
  (ICCV)}, October 2021, pp. 10\,012--10\,022.

\bibitem{canturk2021fuzzy}
{\.I}.~Cant{\"u}rk, ``Fuzzy recurrence plot-based analysis of dynamic and
  static spiral tests of parkinson’s disease patients,'' \emph{Neural
  Computing and Applications}, vol.~33, pp. 349--360, 2021.

\bibitem{khatamino2018deep}
P.~Khatamino, {\.I}.~Cant{\"u}rk, and L.~{\"O}zy{\i}lmaz, ``A deep learning-cnn
  based system for medical diagnosis: an application on parkinson’s disease
  handwriting drawings,'' in \emph{2018 6th International Conference on Control
  Engineering \& Information Technology (CEIT)}.\hskip 1em plus 0.5em minus
  0.4em\relax IEEE, 2018, pp. 1--6.

\bibitem{moro2019forced}
L.~Moro-Velazquez, J.~A. Gomez-Garcia, J.~I. Godino-Llorente, J.~Villalba,
  J.~Rusz, S.~Shattuck-Hufnagel, and N.~Dehak, ``A forced gaussians based
  methodology for the differential evaluation of parkinson's disease by means
  of speech processing,'' \emph{Biomedical Signal Processing and Control},
  vol.~48, pp. 205--220, 2019.

\bibitem{upadhya2018thomson}
S.~S. Upadhya, A.~Cheeran, and J.~H. Nirmal, ``Thomson multitaper mfcc and plp
  voice features for early detection of parkinson disease,'' \emph{Biomedical
  Signal Processing and Control}, vol.~46, pp. 293--301, 2018.

\bibitem{zahid2020spectrogram}
L.~Zahid, M.~Maqsood, M.~Y. Durrani, M.~Bakhtyar, J.~Baber, H.~Jamal,
  I.~Mehmood, and O.-Y. Song, ``A spectrogram-based deep feature assisted
  computer-aided diagnostic system for parkinson’s disease,'' \emph{IEEE
  Access}, vol.~8, pp. 35\,482--35\,495, 2020.

\bibitem{ertuugrul2016detection}
{\"O}.~F. Ertu{\u{g}}rul, Y.~Kaya, R.~Tekin, and M.~N. Almal{\i}, ``Detection
  of parkinson's disease by shifted one dimensional local binary patterns from
  gait,'' \emph{Expert Systems with Applications}, vol.~56, pp. 156--163, 2016.

\bibitem{zhao2022severity}
H.~Zhao, R.~Wang, Y.~Lei, W.-H. Liao, H.~Cao, and J.~Cao, ``Severity level
  diagnosis of parkinson’s disease by ensemble k-nearest neighbor under
  imbalanced data,'' \emph{Expert Systems with Applications}, vol. 189, p.
  116113, 2022.

\bibitem{alzubaidi2021role}
M.~S. Alzubaidi, U.~Shah, H.~Dhia~Zubaydi, K.~Dolaat, A.~A. Abd-Alrazaq,
  A.~Ahmed, and M.~Househ, ``The role of neural network for the detection of
  parkinson’s disease: A scoping review,'' in \emph{Healthcare}, vol.~9,
  no.~6.\hskip 1em plus 0.5em minus 0.4em\relax Multidisciplinary Digital
  Publishing Institute, 2021, p. 740.

\bibitem{aversano2020early}
L.~Aversano, M.~L. Bernardi, M.~Cimitile, and R.~Pecori, ``Early detection of
  parkinson disease using deep neural networks on gait dynamics,'' in
  \emph{2020 International Joint Conference on Neural Networks (IJCNN)}.\hskip
  1em plus 0.5em minus 0.4em\relax IEEE, 2020, pp. 1--8.

\bibitem{Elmaachi}
\BIBentryALTinterwordspacing
I.~E. Maachi, G.~Bilodeau, and W.~Bouachir, ``Deep 1d-convnet for accurate
  parkinson disease detection and severity prediction from gait,'' \emph{CoRR},
  vol. abs/1910.11509, 2019. [Online]. Available:
  \url{http://arxiv.org/abs/1910.11509}
\BIBentrySTDinterwordspacing

\bibitem{hoang2019gait}
N.~S. Hoang, Y.~Cai, C.-W. Lee, Y.~O. Yang, C.-K. Chui, and M.~C.~H. Chua,
  ``Gait classification for parkinson's disease using stacked 2d and 1d
  convolutional neural network,'' in \emph{2019 International Conference on
  Advanced Technologies for Communications (ATC)}.\hskip 1em plus 0.5em minus
  0.4em\relax IEEE, 2019, pp. 44--49.

\bibitem{setiawan2021implementation}
F.~Setiawan and C.-W. Lin, ``Implementation of a deep learning algorithm based
  on vertical ground reaction force time--frequency features for the detection
  and severity classification of parkinson’s disease,'' \emph{Sensors},
  vol.~21, no.~15, p. 5207, 2021.

\bibitem{hochreiter1997long}
S.~Hochreiter and J.~Schmidhuber, ``Long short-term memory,'' \emph{Neural
  computation}, vol.~9, no.~8, pp. 1735--1780, 1997.

\bibitem{Zhao}
A.~Zhao, L.~Qi, J.~Li, J.~Dong, and H.~Yu, ``A hybrid spatio-temporal model for
  detection and severity rating of parkinson’s disease from gait data,''
  \emph{Neurocomputing}, vol. 315, pp. 1--8, 2018.

\bibitem{xia2019dual}
Y.~Xia, Z.~Yao, Q.~Ye, and N.~Cheng, ``A dual-modal attention-enhanced deep
  learning network for quantification of parkinson’s disease
  characteristics,'' \emph{IEEE Transactions on Neural Systems and
  Rehabilitation Engineering}, vol.~28, no.~1, pp. 42--51, 2019.

\bibitem{Gehring}
J.~Gehring, M.~Auli, D.~Grangier, D.~Yarats, and Y.~N. Dauphin, ``Convolutional
  sequence to sequence learning,'' in \emph{International Conference on Machine
  Learning}.\hskip 1em plus 0.5em minus 0.4em\relax PMLR, 2017, pp. 1243--1252.

\bibitem{klambauer2017self}
G.~Klambauer, T.~Unterthiner, A.~Mayr, and S.~Hochreiter, ``Self-normalizing
  neural networks,'' in \emph{Proceedings of the 31st international conference
  on neural information processing systems}, 2017, pp. 972--981.

\bibitem{goldberger2000physiobank}
A.~L. Goldberger, L.~A. Amaral, L.~Glass, J.~M. Hausdorff, P.~C. Ivanov, R.~G.
  Mark, J.~E. Mietus, G.~B. Moody, C.-K. Peng, and H.~E. Stanley, ``Physiobank,
  physiotoolkit, and physionet: components of a new research resource for
  complex physiologic signals,'' \emph{circulation}, vol. 101, no.~23, pp.
  e215--e220, 2000.

\bibitem{frenkel2005effect}
S.~Frenkel-Toledo, N.~Giladi, C.~Peretz, T.~Herman, L.~Gruendlinger, and J.~M.
  Hausdorff, ``Effect of gait speed on gait rhythmicity in parkinson's disease:
  variability of stride time and swing time respond differently,''
  \emph{Journal of neuroengineering and rehabilitation}, vol.~2, no.~1, p.~23,
  2005.

\bibitem{frenkel2005treadmill}
------, ``Treadmill walking as an external pacemaker to improve gait rhythm and
  stability in parkinson's disease,'' \emph{Movement disorders: official
  journal of the Movement Disorder Society}, vol.~20, no.~9, pp. 1109--1114,
  2005.

\bibitem{yogev2005dual}
G.~Yogev, N.~Giladi, C.~Peretz, S.~Springer, E.~S. Simon, and J.~M. Hausdorff,
  ``Dual tasking, gait rhythmicity, and parkinson's disease: which aspects of
  gait are attention demanding?'' \emph{European journal of neuroscience},
  vol.~22, no.~5, pp. 1248--1256, 2005.

\bibitem{hausdorff2007rhythmic}
J.~M. Hausdorff, J.~Lowenthal, T.~Herman, L.~Gruendlinger, C.~Peretz, and
  N.~Giladi, ``Rhythmic auditory stimulation modulates gait variability in
  parkinson's disease,'' \emph{European Journal of Neuroscience}, vol.~26,
  no.~8, pp. 2369--2375, 2007.

\bibitem{bertasius2021SpaceTimeAttention}
G.~Bertasius, H.~Wang, and L.~Torresani, ``Is {Space}-{Time} {Attention} {All}
  {You} {Need} for {Video} {Understanding}?'' in \emph{{ICLR}}, 2021.

\end{thebibliography}

\end{document}